\documentclass[lettersize,journal]{IEEEtran}
\usepackage{amsmath,amsfonts}
\usepackage{algorithmic}
\usepackage{algorithm}
\usepackage{array}
\usepackage[caption=false,font=normalsize,labelfont=sf,textfont=sf]{subfig}
\usepackage{textcomp}
\usepackage{stfloats}
\usepackage{url}
\usepackage{verbatim}
\usepackage{graphicx}
\usepackage{cite}
\usepackage{subfig}
\usepackage[export]{adjustbox}
\usepackage{hyperref}

\begin{document}

\title{Towards Continual, Online, Self-Supervised Depth}

\author{Muhammad Umar Karim Khan
\thanks{Muhammad Umar Karim Khan is with ISize Ltd, London, UK \{email:umar@isize.co\}.}
\thanks{Manuscript received April XX, XXXX;}}

\markboth{IEEE Trans. on Circuits and Systems for Video Technology,~Vol.~XX, No.~XX, August~XXXX}%
{Shell \MakeLowercase{\textit{Khan, Muhammad Umar Karim}}: Towards Continual, Online, Self-Supervised Depth}

\IEEEpubid{0000--0000/00\$00.00~\copyright~2021 IEEE}

\maketitle


\begin{abstract}
Although depth extraction with passive sensors has seen remarkable improvement with deep learning, these approaches may fail to obtain correct depth if they are exposed to environments not observed during training. Online adaptation, where the neural network trains while deployed, with self-supervised learning provides a convenient solution as the network can learn from the scene where it is deployed without external supervision. However, online adaptation causes a neural network to forget the past. Thus, past training is wasted and the network is not able to provide good results if it observes past scenes. This work deals with practical online-adaptation where the input is online and temporally-correlated, and training is completely self-supervised. Regularization and replay-based methods without task boundaries are proposed to avoid catastrophic forgetting while adapting to online data. Effort has been made to make the proposed approach suitable for practical use. We apply our method to both structure-from-motion and stereo depth estimation. We evaluate our method on diverse public datasets that include outdoor, indoor and synthetic scenes. Qualitative and quantitative results with both structure-from-motion and stereo show superior forgetting as well as adaptation performance compared to recent methods. Furthermore, the proposed method incurs negligible overhead compared to fine-tuning for online adaptation, proving to be an adequate choice in terms of plasticity, stability and applicability. The proposed approach is more inline with the artificial general intelligence paradigm as the neural network learns continually with no supervision. Source code is available at \texttt{https://github.com/umarKarim/cou\_sfm} and \texttt{https://github.com/umarKarim/cou\_stereo}.
\end{abstract}


\section{Introduction}
\IEEEPARstart{D}{eep} learning mimics the human behavior from certain perspectives, though much is yet to be achieved. The results of deep learning reflect our behavior much better compared to classical algorithms, with Catastrophic forgetting \cite{kirkpatrick2017overcoming} and online adaptation being two specific examples. Neural networks recall recent information much better compared to the distant past. Similarly, neural networks require special mechanisms to retain past information. 
 
 Extracting depth of a scene is critical for scene perception. Depth serves a wide variety of applications from computational photography to navigation. Active sensors such as LIDAR or time-of-flight are capable of providing results at depth resolutions better than the human vision system. However, these have numerous challenges such as low spatial resolution, interference, and high power consumption. Efforts have been made towards passive sensors replicating the human vision system. For example, stereo cameras, structure-from-motion (SfM) and monocular methods replicate binocular vision, head motion and object-size information for depth extraction, respectively. 
 
 Despite the widely-acclaimed success of deep learning, generalization towards unseen samples has proven to be a major problem. This problem is usually tackled by training over a dataset that covers as much of the possible samples as possible. For example, MiDaS \cite{ranftl2019towards} adopts this strategy and produces great results for a wide variety of scenes. However, the possibility of an edge case still exists, which can prove to be catastrophic. Furthermore, any small addition to the training data requires performing retraining over the whole dataset again. Thus, despite deep learning approaches outperforming classical methods in depth extraction, their generalization to unseen samples is a challenge yet to be resolved. 
 
 A different approach towards generalization is to allow the neural network to be always in an active, learning state. In other words, the neural network learns about the scene where it is deployed while not forgetting past information. This approach not only allows generalization to new domains but also does not require retraining. Thus, edge devices can autonomously learn without being connected to a centralized system. 
 
 The above paradigm shift comes with numerous challenges. Directly learning new tasks causes catastrophic forgetting of past tasks. Thus, the performance may improve on the new domain but the system will have relearn once it is used in the past domain. Also, in conventional training, samples are passed as a batch and repeatedly used with shuffling to train the neural network. However, in the above setting, only one sample is available at a time and that cannot be used repeatedly to train the neural network. Furthermore, task boundaries are not available, i.e., there is no supervision about when a new scene starts. Finally, it is not practical to provide labels for training for every or any sample.
 
 Numerous notable approaches have been proposed for the above challenges separately. These include continual, online, task-free continual, and self-supervised learning. However, little effort has been made towards solutions that can tackle all these problems together. Furthermore, evaluations have been limited to small and simple datasets in general. 
 
 \begin{figure*}[!hbt]
    \small
    \begin{center}
       \includegraphics[width=0.8\textwidth]{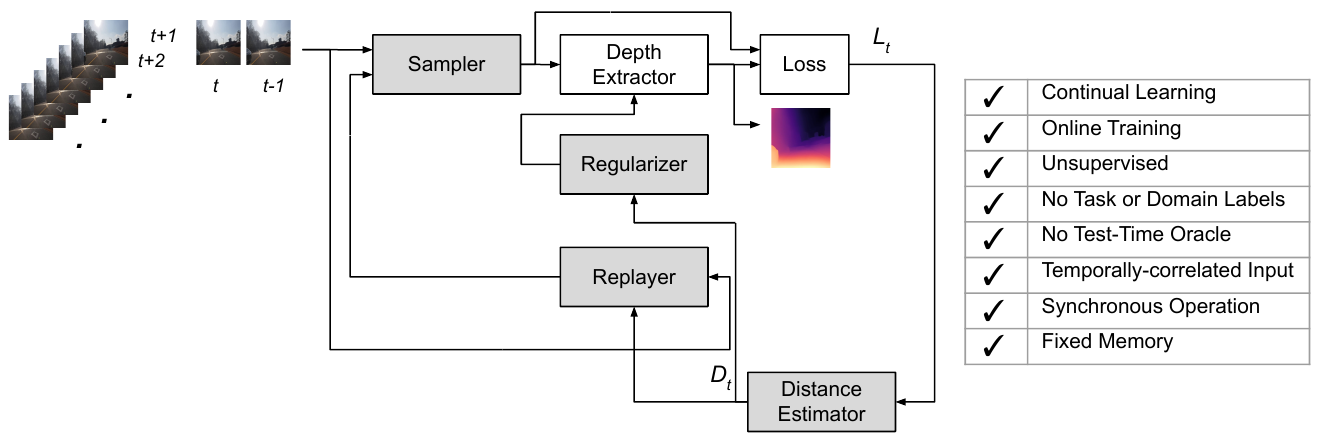}
       \caption{A graphical description of the proposed method. The non-shaded blocks are from conventional self-supervised depth-extraction schemes. The shaded blocks allow continual and online operation.}
       \label{fig:intro_fig}
    \end{center}
 \end{figure*}

 This paper aims towards providing a unified solution to the above challenges for depth extraction. Our key contributions are as follows.
 \begin{itemize}
    \item A novel regularization and replay-based method for online learning of depth while avoiding catastrophic forgetting is proposed.
    \item Practical constraints such as unsupervised learning, no task (scene) labels, no test-time oracle (where task information is available at test time), temporally-correlated input, synchronous operation and fixed memory are considered.
    \item Experiments are conducted on both SfM and stereo, which show the plasticity (online adaptation) and stability (forgetting) performance under the same settings on challenging datasets.
 \end{itemize} 
 
 Although this work is limited to depth extraction, the principles proposed here can be used in any downstream self-supervised task. Successful implementation of the proposed method on edge devices provides a step closer to artificial general intelligence, where devices can learn and make decisions on their own while deployed.
 
 The rest of the paper is structured as follows. Section II describes the related work in depth estimation, continual and online learning, and continual learning methods for depth estimation. Section III presents the proposed method for soft task-boundaries. The proposed regularization and replay approaches are discussed in Sections IV, V, and VI. Our choice of neural networks is discussed in Section VII. Section VIII details the experimental results and the paper is concluded in Section IX.
 
 \section{Related Work}
 \subsection{Depth Estimation}
\noindent  Self-supervised training for stereo-based depth estimation is based on reconstructing one stereo image (or depth map) from the other \cite{garg2016unsupervised, godard2017unsupervised}. Similar approaches have been proposed for monocular or SfM approaches \cite{zhou2017unsupervised, godard2019digging}, where the next or previous frame is reconstructed using depth and pose estimated by neural networks. \cite{gonzalezbello2020forget} proposes excluding the occluded regions in training. Wasserstein distance is used in \cite{garg2020wasserstein} for depth estimation and 3D object detection. Authors in \cite{hur2020self} propose a self-supervised scheme to estimate the depth and 3D scene flow simultaneously. Multi-camera systems have been proposed in \cite{zhang2020depth,imran2020unsupervised} to increase depth range. \cite{poggi2020uncertainty} proposes a method to measure confidence of self-supervised depth estimation. In \cite{watson2019self}, authors use classical stereo approaches to improve performance of monocular depth estimation schemes. In \cite{spencer2020defeat}, authors propose features for self-supervised depth estimation that are robust to the changes in domain. SLAM and monocular depth estimation are used together to improve each other in \cite{tiwari2020pseudo}. In \cite{badki2020bi3d}, stereo depth is estimated by binary classification.  Removing camera rotations \cite{zhou2019moving} and textured patches \cite{yu2020p} have been used for depth estimation of indoor scenes. Authors in \cite{wu2021toward} distill information from a network which only provides relative depth of objects in a scene for depth estimation of indoor scenes. \cite{walia2022gated2gated} provides a method for self-supervised depth estimation from gated images. DiffNet \cite{zhou2021self} uses the HRNet \cite{wang2020deep} proposed for the task of semantic segmentation for self-supervised depth estimation and shows superior performance on the KITTI \cite{Menze2015CVPR} dataset.
 
 \subsection{Continual and Online Learning}
\noindent In continual learning, a neural network learns new tasks while not forgetting about the old ones. Broadly, these methods use distillation \cite{li2017learning, fini2020online}, regularization \cite{kirkpatrick2017overcoming, aljundi2018memory}, replay \cite{shin2017continual, gupta2020maml} or their combination \cite{pan2020continual}. Expansion-based methods can be considered special form of regularization. Regularization-based methods penalize changes to important weights or add new computational nodes to the neural network with a new task. Replay-based schemes feed past data to neural networks while learning the new task. The effect of different training regimes on catastrophic forgetting has been analyzed in \cite{mirzadeh2020understanding}. \cite{buzzega2020dark} proposes a continual learning method with constant memory, no test-time oracle and without task boundaries. In \cite{caccia2020online}, fast adaptation and recall are achieved for continual learning. \cite{chen2020mitigating} proposes to determine the instance of an input sample and use it to train the network accordingly. Knowledge transfer in similar tasks is dealt with in \cite{ke2020continual}. \cite{he2020incremental} deals with continual learning where new tasks may contain samples from old tasks. \cite{abati2020conditional} gates layers based on the current task and also predicts tasks at inference. \cite{wang2020visual} proposes an online learning method for detecting interesting scenes for mobile robots. \cite{chrysakis2020online} proposes a method to perform continual learning with temporally-correlated data streams and \cite{aljundi2019task} uses the loss curve to define task boundaries. \cite{hayes2020remind} and \cite{pellegrini2020latent} use network features to partially update the neural network for fast operation.
 
 Although approaches have been proposed for continual learning with replay under numerous constraints, there are not methods that satisfy all the constraints mentioned in this work. Authors of \cite{aljundi2019online} propose a method of continual and online learning by using samples for replay that are supposed to show the worst performance if the network is updated with the current online samples. They estimate the loss for numerous past samples with each online batch (up to 50 for MNIST). In other words, they require multiple forward passes while online training. This may not be a problem with small networks but is not feasible for a practical online training network. \cite{aljundi2019gradient} also proposes a replay-based approach where the replay buffer is filled with samples such that the replay samples are as varied as possible. However, their method is also not feasible for real-time operation because it requires computing the dot-product between the gradients of the online batch and at least a subset of the replay buffer with every online batch. Authors of \cite{javed2019meta} divide the network into Representation Learning and Prediction Learning sub-networks. They train such that the features learned by Representation Learning allow forward transfer and avoid forgetting for online batches. Their method, however, would also require multiple forward and backward passes in an inner loop for a single batch of online data. \cite{rao2019continual} and \cite{lee2019neural} are methods for task-free continual learning. These methods are based on expanding the neural network. In practise, the number of tasks is unknown beforehand. In other words, there is no limit on the computational complexity of these systems, making them less feasible for practical applications.

 \subsection{Continual and Online Learning of Depth}
\noindent Numerous authors have worked towards the domain adaptation problem for depth estimation. In \cite{zhang2020depth} and \cite{zheng2018t2net}, authors propose a method to adapt from synthetic datasets to real datasets. \cite{sun2022learn} proposed an adversarial approach for online adaptation across datasets. In \cite{tonioni2017unsupervised}, the authors propose using conventional stereo matching algorithms for domain adaptation. To speed up the process, authors propose modular adaptation in \cite{tonioni2019real}. Authors in \cite{tonioni2019learning} propose a meta-learning objective to quickly adapt to new scenes. Their work has been advanced in \cite{zhang2020online}, where the authors develop an approach to perform online adaptation without catastrophic forgetting. \cite{kuznietsov2021comoda} uses replay at test time.  \cite{vodisch2022continual} proposes using two networks; one for adapting to the current scene while the other to generatlize over past scenes. Note that \cite{tonioni2019learning}, \cite{zhang2020online} and \cite{kuznietsov2021comoda} perform adaptation on test data. Furthermore, \cite{tonioni2019real} resets the neural network for every test sequence. 
 
 \subsection{This Work}
\noindent Methods have been proposed that deal with the individual challenges listed in Fig. \ref{fig:intro_fig}, however, we could not find methods which deal with all these challenges together. Methods that continually learn depth either present results of forgetting or adaptation in separate experiments, thus, the plasticity-stability compromise cannot be fully understood. Generally, similar datasets are used to evaluate online adaptation from one dataset to another and test data adaptation is performed. In this work, both adaptation and forgetting are analyzed over different datasets without adaptation over the test data. 
 
 \section{Soft Task-Boundaries}
 \noindent Unlike general continual learning research, the task boundaries are unknown, which is more practical. Different scenes can be treated as different tasks. For example, day-time images represent a different task compared to night images. Similarly, indoor scenes represent a different task compared to outdoor scenes. Furthermore, a task can be further sub-divided into sub-tasks. For example, indoor scenes can be divided into classroom and kitchen scenes, and outdoor scenes can be divided into residential and highway scenes. The proposed approach is applicable to any classification of tasks.
 
 The profile of the loss can provide us with valuable hints towards task boundaries. A converged loss is expected to remain stable over a period of time. A significant change in the loss value may indicate that the task has changed. 
 
 Let $L^{(t)}$ be a random variable indicating the loss value at time $t$. It is assumed 
 
 \begin{equation}
    L^{(t)} \sim \mathcal{N}(\mu_t, \sigma_t),
 \end{equation}
 
 \noindent with $\mu_t$ and $\sigma_t$ the corresponding mean and standard deviation, respectively. It can be derived (see Appendix)
 
 \begin{equation}
   -\log P(T(t) = T(t - 1)) = \frac{1}{2}D_t(L^{(t - 1)}, L_{t}(x_t; \theta_{t})) + c,
 \end{equation}
 
 \noindent where $T(t)$ is the task at time $t$, $c = \log(\sqrt{2\pi}\sigma_{t-1})$, $\theta_t$ represent the network parameters and $D_t$ is the squared Mahalanobis distance given by 
 
 \begin{equation}
    D_t(L^{(t-1)}, L_{t}(x_t; \theta_{t})) = \frac{(L_{t}(x_t; \theta_{t}) - \mu_{t-1}) ^ 2}{\sigma_{t-1}^2}.
 \end{equation}
  
 \noindent In detail, the log-probability of a new task at time $t$ depends on how far the current observed loss $L_{t}$ is from the loss distribution of $L^{(t-1)}$. The parameters are updated iteratively as
 \begin{equation}
    \mu_{t} = \mu_{t-1} + \alpha_L ( L_{t} - \mu_{t-1})
 \end{equation} 
 
 \noindent and 
 
 \begin{equation}
    \sigma_{t}^2 = \sigma_{t-1}^2 + \alpha_L \left((L_{t} - \mu_{t})^2 - \sigma_{t-1}^2\right),
 \end{equation}
 
 \noindent where $\alpha_L$ is a constant \cite{stauffer1999adaptive}. The loss profile has been used to define task boundaries in the past as well \cite{aljundi2019task}. However,  \cite{aljundi2019task} uses plateus and peaks of the loss profile to determine a task boundary whereas here just the Mahalanobis distance is used. Furthermore, they use hard task-boundaries whereas here the Mahalanobis distance is used as a representation of the probability of the task boundary. In other words, soft task-boundaries are used here.
 
 \section{Regularization with Soft Task-Boundaries}
 \noindent Regularization for continual learning limits changes to the network architecture or parameters across tasks to retain information about previous tasks. Memory-aware synapses (MAS) \cite{aljundi2018memory} is a known approach, which assigns importance to parameters based on how changing them changes the output for a given input. A sample is passed through the network at the beginning of every task and the gradient determines the importance of parameters. This creates numerous problems. First, hard task-boundaries are required. Second, an additional forward-backward pass is performed at the end of each task.
 
 In this work, inspiration is taken from neural compression techniques, where parameters close to zero are compressed \cite{li2016pruning}. This is quite intuitive as a parameter with larger magnitude will affect the output more than a parameters with smaller magnitude. Thus, the importance of the $i$-th parameter is given by 
 
 \begin{equation}
    \Omega_{i, t} = |\theta_{i, t-1}|,
 \end{equation}
 
 \noindent where $\theta_i$ is the $i$-th parameter of the network. The regularization loss then becomes 
 
 \begin{equation}
    L_t^{(reg)}(\theta) = \sum_i  \Omega_{i, t} |\theta_i - \theta_{i, t -1}|, 
 \end{equation}
 
 \noindent and the combined loss is
 
 \begin{equation}
    L_t^{(tot)}(x_t, \theta) = L_t(x_t; \theta) + \gamma D_t L_t^{(reg)}(\theta),
 \end{equation}
 
 \noindent where $\gamma$ is a constant. In detail, the method is more conservative about changing the important parameters if the chances of a new task are higher. The proposed approach assists in both remembering past tasks as well as forward transfer, does not require hard task boundaries and does not require additional forward-backward passes. Note that $\theta_t$ represents the current value of the parameters whereas $\theta$ represents the decision variable. 
 
 \section{Replay with Soft Task-Boundaries}
 \noindent Let $T$ represent the current task. A general representation of the overall loss function for the task $T$ with regularization is  
 
 \begin{equation} 
    L_T^{(ov)} = L_T^{(curr)} + \gamma L_T^{(prev)}.
 \end{equation}
 
 \noindent Minimizing $L_T^{(curr)}$ and $L_T^{(prev)}$ is equivalent to improving performance on the current and previous tasks, respectively. The loss on the previous task can be similarly given as 
 
 \begin{equation}
    L_T^{(prev)} \equiv L_{T-1}^{(ov)} = L_{T-1}^{(curr)} + \gamma L_{T-1}^{(prev)}.
 \end{equation}
 
 \noindent Thus, the contribution of the $n$-th previous task to the current loss function is $\gamma^n$. Since $\gamma$ is generally set to less than one, it is expected that the performance over distant tasks will be poorer compared to recent tasks.
 
 To resolve this issue, we use sample replay which assigns equal importance to all past tasks. Samples are stored in the replay memory if $D_t$ is greater than one. By using this approach, difficult samples are saved from which the network can learn more \cite{chrysakis2020online}. The storage capacity is limited to 1.5GB \cite{hayes2020remind}. If at full capacity, the new sample replaces a randomly-chosen old sample. 
 
 The choice of storing original images to memory is based on numerous factors. Replay with a generative network \cite{shin2017continual} is an option. For classification problems, generative replay may work well as classification is based on abstract features. Depth extraction, on the other hand, is based on pixel-level disparities which are difficult to recreate. Furthermore, generative replay requires significant amount of additional computational power. Some authors \cite{hayes2020remind} propose storing compressed intermediate features and their corresponding target labels for partially updating the neural network. However, storing target labels is not possible here. First, such an approach requires storing the depth maps as targets. Second, the generated depth maps may not be accurate enough to be used as targets. Replay with SfM has been used in \cite{kuznietsov2021comoda}, however, their method is solely focused on adaptation; they do not limit replay memory; they adapt on test data; and they do not update the replay buffer while training online. The proposed method, on the other hand, focuses on both adaptation and forgetting; has limited replay memory; does not adapt on test data; and updates the replay buffer while training online. The last point is important as it allows the proposed method to remember depth of older scenes observed during online training. 
 
 The replay and online samples are input to the neural network with the same probability. By this the paradigm of spaced repetition \cite{amiri2017repeat} is followed where difficult examples are taught to a learning system spaced over time. 
 
 \section{Replay vs Regularization}
  \noindent Regularization requires additional memory and computations to maintain the importance of the parameters. Replay does not require significant amount of computational power, however, it needs space to store past samples. Therefore, replay can be used when sufficient storage is available and read/write is fast. Both the proposed regularization and replay are used together in this work.
 
 \section{Self-Supervised Depth Extraction}
 \begin{figure}[t]
    \begin{center}
       \includegraphics[width=0.8\columnwidth]{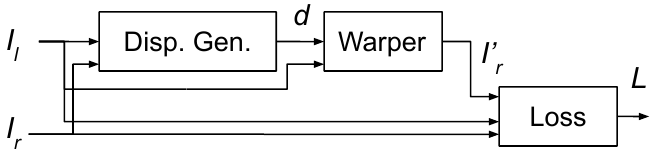}
       \caption{Top-level block diagram of stereo-based disparity estimation where Disparity Generator is a neural network with learnable parameters.}
       \label{fig:block_stereo}
    \end{center}
 \end{figure}
 
 \begin{figure}[t]
 \begin{center}
    \includegraphics[width=0.9\columnwidth]{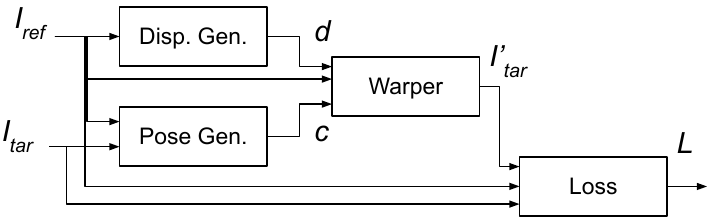}
    \caption{Top-level block diagram of SfM-based disparity and pose estimation where Disparity Generator and Pose Generator are neural networks with learnable parameters.}
    \label{fig:block_sfm}
 \end{center}
 \end{figure}
 
 \begin{figure*}[t]
 \begin{center}
    \includegraphics[width=0.8\textwidth]{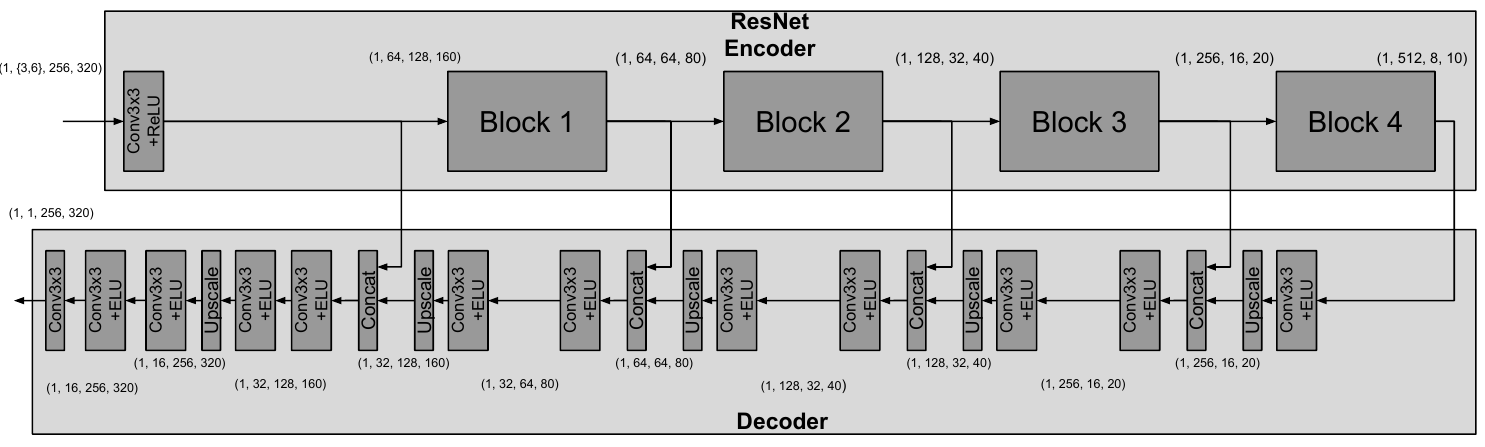}
    \caption{The Disparity Generator network used in stereo and SfM, where feature dimensions are shown as NCHW. The input has three or six channels depending on whether SfM or stereo is used. Refer to \cite{he2016deep} for the ResNet Encoder.}
    \label{fig:disp_ops}
 \end{center}
 \end{figure*}
 
 \begin{figure*}[t]
 \begin{center}
    \includegraphics[width=0.8\textwidth]{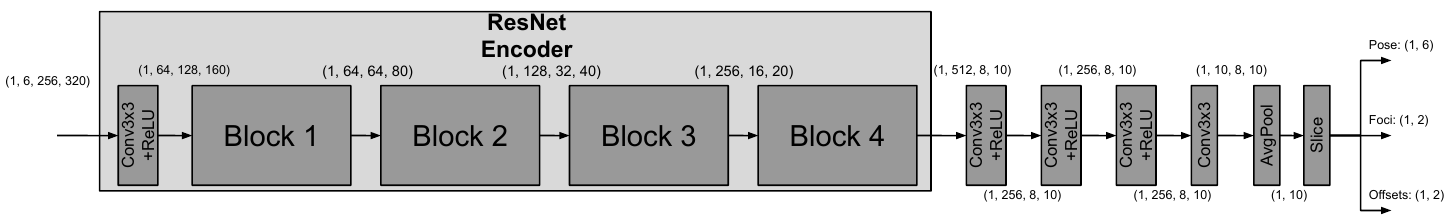}
    \caption{The Pose Generator network used in SfM, where feature dimensions are shown as NCHW. Refer to \cite{he2016deep} for the ResNet Encoder.}
    \label{fig:pose_ops}
 \end{center}
 \end{figure*}
 
 \noindent The proposed approaches are applied to both self-supervised stereo and SfM. 
 In the following, the subscript $t$ is skipped for brevity.
 
 \subsection{Self-Supervised Stereo}
 Depth extraction with stereo cameras is based on disparity estimation across rectified left and right images. The relationship between left and right images is given by 
 
 \begin{equation}
    I_l(x + d(x, y), y) = I_r(x, y),
 \end{equation}
 
 \noindent where $I_l$, $I_r$, and $d$ denote the left image, right image, and disparity map, respectively. A network is trained to implicitly learn the disparity by reconstructing the right image. I.e., 
 
 \begin{equation}
    d = \mathcal{F}_d(I_l, I_r; \theta),
 \end{equation}
 
 \noindent which reconstructs the right image as 
 
 \begin{equation}
    I'_r = f_w(I_l; d),
 \end{equation}
 
 \noindent where $\mathcal{F}_d$ represents the disparity neural network and $f_w$ represents the warping function followed by bilinear interpolation \cite{jaderberg2015spatial}. The reconstruction loss is then given by 
 
 \begin{equation}\label{eq:loss_st}
    L = L_{p}(I'_r, I_r) + \beta_{ss} L_{ss}(I'_r, I_r) + \beta_s L_{s}(d, I_l),
 \end{equation}
 
 \noindent where $L_{p}$, $L_{ss}$, and $L_{s}$ are pixel-wise, SSIM and smoothness losses, respectively. $\beta_{ssim}$ and $\beta_{s}$ are constants. Refer to \cite{godard2017unsupervised,godard2019digging} for a discussion on these constants and losses. The approach is shown in Fig. \ref{fig:block_stereo}.  
 
 Numerous techniques have been proposed to improve the performance of self-supervised stereo. For exampler, \cite{godard2017unsupervised} proposes to reconstruct both the left and right images, and \cite{pilzer2018unsupervised} uses cycle GANs. Since in this work the emphasis is on online operation, only the right image is reconstructed and GANs are not used.
 
 \subsection{Self-Supervised Structure-from-Motion}
 For estimating depth using SfM, two separate networks are used. The first generates the disparity map while the second generates the camera matrix with two frames as input, one as reference and the other as target. The target frame is reconstructed from the reference frame, disparity and camera matrix using the warping function \cite{jaderberg2015spatial}
 
 \begin{equation}
    I'_{tar} = f_w^{(c)}(I_{ref}; d, c),
 \end{equation}
 
 \noindent where $f_w^{(c)}$ represents the warping function \cite{jaderberg2015spatial}, $d$ is the disparity, and $c$ is the camera matrix. The approach is shown in Fig. \ref{fig:block_sfm}. The current frame and previous frame are the target and reference, and vice versa. Eq. \ref{eq:loss_st} is used to compute the loss between the reconstructed and original target frame. Geometric loss is used as in \cite{bian2019unsupervised}. 
 
 \subsection{Network Architecture}
 The network architecture for the disparity and pose generators are shown in Fig. \ref{fig:disp_ops} and Fig. \ref{fig:pose_ops}, respectively. The disparity generator has an archictecture similar to the UNet \cite{ronneberger2015u} with an encoder and decoder. More specifically, the disparity generator is composed of a ResNet18 encoder \cite{he2016deep} and a convolutional decoder. The network architecture is similar to the networks used in the past for monocular depth-estimation \cite{godard2017unsupervised}, \cite{godard2019digging}, \cite{bian2019unsupervised}. There are a couple of key differences. For stereo disparity estimation, we concatenate the left and the right images before passing them to the disparity generator network. Furthermore, we use a relatively small image resolution of $320\times256$ to reduce computational complexity. The pose generator also includes the ResNet18 encoder followed by multiple convolutional layers.  However, unlike \cite{bian2019unsupervised} and majority of SfM-based deep learning approaches, the pose network learns both the camera intrinsics and extrinsics \cite{gordon2019depth}. This is more practical as the intrinsics are not obtained beforehand and the method is camera agnostic. In the rest of the paper, these networks with the proposed subtle changes will be termed as Proposed.

 \begin{table}
    \caption{Absolute Relative of depth estimation. \textit{FT}, \textit{K}, \textit{N} and \textit{vK} stand for fine tuning, KITTI, NYU and virtual KITTI, respectively. Best is bold and second best is underlined.}
    \label{tab:sfm_stereo}
    \begin{center}
       \small
       \begin{tabular}{lll|cc|cc}
          \hline 
          \multicolumn{3}{c}{\textbf{Mode}} & \multicolumn{2}{c}{\textbf{SfM}} & \multicolumn{2}{c}{\textbf{Stereo}} \\
          \hline
          \begin{tabular}[x]{@{}c@{}}Online\\Train\end{tabular} 
           & Net. & Meth. & TDP & NDP & TDP & NDP\\ 
          \hline \hline
          K & \cite{zhou2021self} & FT & 0.2214 & 0.5746 & 0.1932 & 0.1996 \\
          K & \cite{zhou2021self} & \cite{kuznietsov2021comoda} & 0.1904 & 0.2265 & 0.2098 & \underline{0.1670}\\
          K & \cite{zhou2021self} & Prop. & 0.1916 & 0.2694 & 0.2123 & 0.1902\\ 
          K & Prop. & FT & 0.1895 & 0.3504 & \underline{0.1920} & 0.1980\\
          K & Prop. & \cite{kuznietsov2021comoda} & \underline{0.1580} & \textbf{0.1911} & 0.1962 & 0.1685\\
          K & Prop. & Prop. & \textbf{0.1543} & \underline{0.1952} & \textbf{0.1825} & \textbf{0.1660} \\ 
          \hline 
          N,vK & \cite{zhou2021self} & FT & 0.3243 & 0.3088 & 0.1860 & 0.2401 \\
          N,vK & \cite{zhou2021self}& \cite{kuznietsov2021comoda} & 0.2243 & 0.1695 & 0.1803 & 1.4442\\
          N,vK & \cite{zhou2021self} & Prop. & 0.2379 &  0.1995  & \underline{0.1765} & 0.2190 \\ 
          N,vK & Prop. & FT & 0.2430 & 0.3336 & 0.1991 & \underline{0.2090}\\
          N,vK & Prop.& \cite{kuznietsov2021comoda} & \underline{0.1912} & \textbf{0.1586}  & 0.1834 & 0.3177\\
          N,vK & Prop.& Prop. & \textbf{0.1872} & \underline{0.1624} & \textbf{0.1653} & \textbf{0.1770}\\ 
          \hline
       \end{tabular}
    \end{center}
    
 \end{table}

 \section{Experimental Results}
 \noindent Since the proposed approach is for online learning and not based on the conventional approach of training followed by testing, therefore, some terminology and evaluation protocols are discussed first.
 
 \subsection{Datasets}
 \noindent For experiments, three datasets are used: KITTI \cite{Menze2015CVPR}, Virtual KITTI \cite{cabon2020vkitti2}, and rectified New York University (NYU) v2 \cite{silberman2012indoor,bian2020unsupervised} datasets. The Eigen test split \cite{eigen2014depth} is used with KITTI. For the Virtual-KITTI dataset, the first 90\% frames of every sequence are used for training and the last 10\% are for evaluation. Refer to supplementary material for more details. These datasets are very different from each other, thereby allowing better evaluation of the proposed approach.
 
 \subsection{Terminology}
 \noindent Both the plasticity (adaptation) and stability (remembrance) performance of the proposed method are evaluated at both the dataset and scene levels. Training Dataset Performance (TDP) shows the results on the dataset over which online training has been performed, whereas Non-training Dataset Performance (NDP) shows the results on the other dataset. TDP shows how well the method is adapting to new scenes as well as how well the network remembers the scenes previously observed during online training. NDP shows how well the network remembers the data observed during pre-training.

 \begin{table}
 \caption{Results of training with KITTI and then with virtual KITTI. *shows that \texttt{fog} and \texttt{rain} are included in online training. Best and second best are bold and underlined, respectively.}
    \label{tab:sfm_kitti_vkitti}
    \begin{center}
       \small
       \begin{tabular}{lccccc}
          \hline 
          Method & Model & RMSE $\downarrow$ & \begin{tabular}[x]{@{}c@{}}Abs.\\Rel.\end{tabular}$\downarrow$ & \begin{tabular}[x]{@{}c@{}}$\delta$\\$(1.25)$\end{tabular}$\uparrow$ & \begin{tabular}[x]{@{}c@{}}$\delta$\\$(1.25^2)$\end{tabular}$\uparrow$ \\
          \hline \hline 
          FT & \cite{zhou2017unsupervised} & 6.5248 & 0.2070 & 0.7041 & 0.8806 \\
          FT & \cite{bian2019unsupervised} & 5.6528 & 0.1735 & 0.7743 & 0.9140 \\
          L2A \cite{tonioni2019learning} & \cite{zhou2017unsupervised} & 6.3804 & 0.1937 & 0.7221 & 0.8980\\
          L2A  \cite{tonioni2019learning}& \cite{bian2019unsupervised} & 5.5500 & 0.1692 & 0.7881 & 0.9197 \\
          LPF  \cite{zhang2020online}& \cite{zhou2017unsupervised} & 6.1090 & 0.1794 & 0.7307 & 0.9126 \\
          LPF  \cite{zhang2020online} & \cite{bian2019unsupervised} & \underline{5.4452}  & 0.1505 & 0.7990 & 0.9325\\
          \hline
          FT & Prop. & 7.4267 & 0.2575 & 0.6106 & 0.8441\\
          Prop. & Prop. & 5.4739 & \textbf{0.1313} & \textbf{0.8345} & \textbf{0.9455} \\
          FT* & Prop. & 7.147 & 0.2158 & 0.6690 & 0.8679 \\
          Prop.* & Prop.  & \textbf{5.3554} & \underline{0.1392} & \underline{0.8251} & \underline{0.9433} \\
          \hline 
       \end{tabular}
    \end{center}
    
 \end{table}

 \begin{figure*}
 \centering
 \subfloat[]{\includegraphics[width=0.5\columnwidth, valign=t]{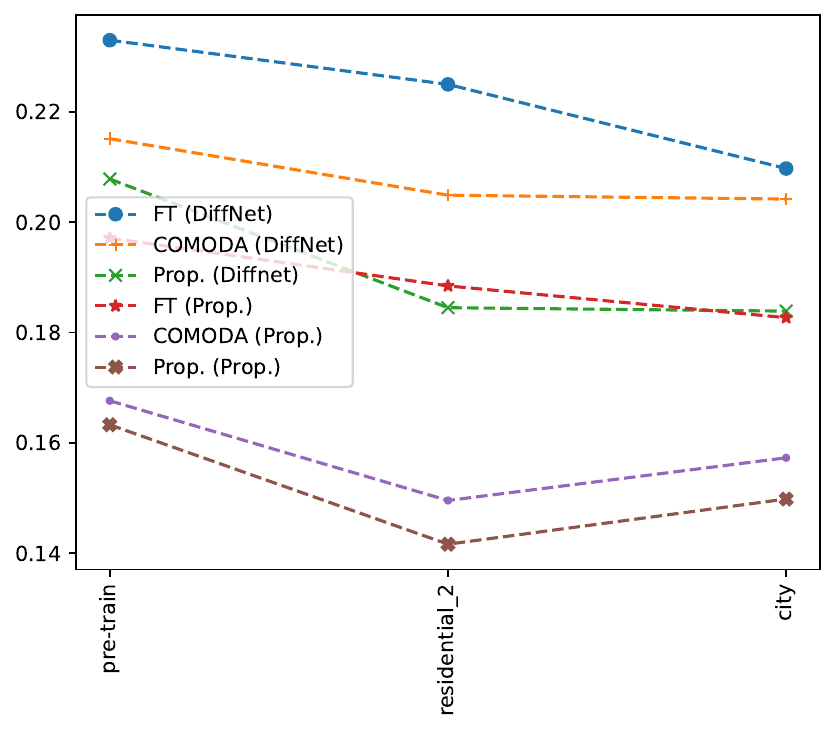}} \hfill
 \subfloat[]{\includegraphics[width=0.5\columnwidth, valign=t]{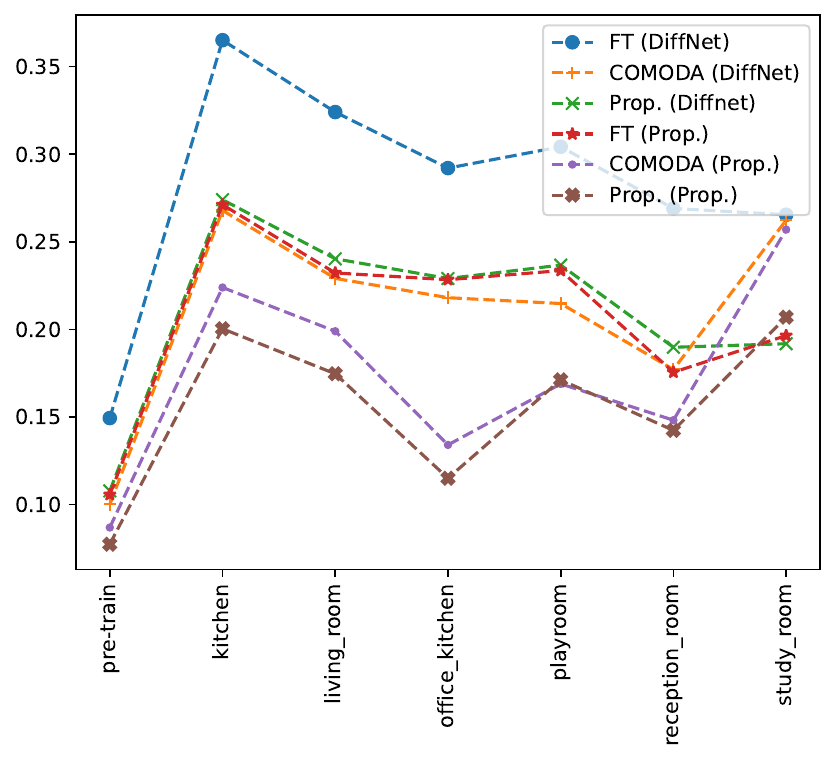}} \hfill
 \subfloat[]{\includegraphics[width=0.5\columnwidth, valign=t]{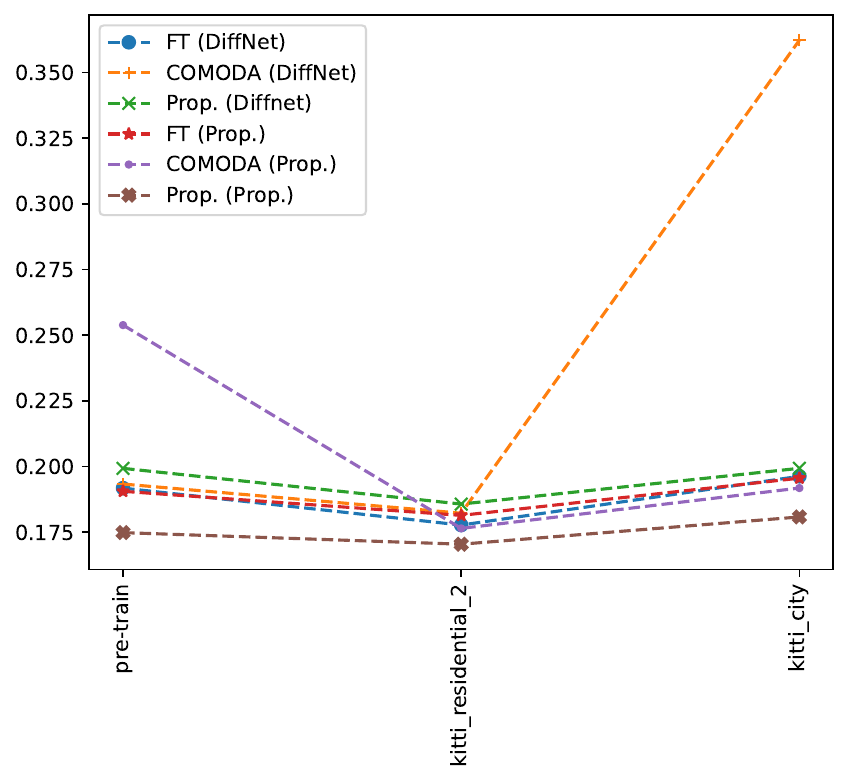}} \hfill
 \subfloat[]{\includegraphics[width=0.5\columnwidth, valign=t]{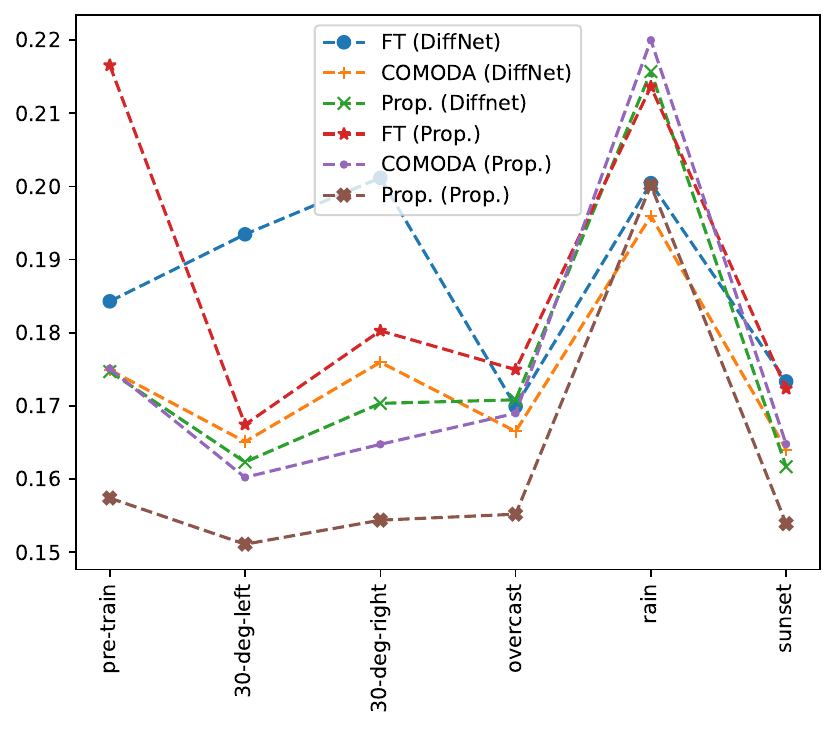}} \hfill
\caption{Results of the absolute-relative metric on SfM-based training on individual scenes of training data where the training data and scenes are from (a) KITTI and (b) NYU datasets. Similarly, results of stereo-based training on individual scenes of training data where the training data and scenes are from (c) KITTI and (d) virtual KITTI. Results are named as \textit{Continual learning method (Network)}.}
\label{fig:ind_scenes}
\end{figure*}

 \begin{figure*}[!ht]
    \begin{center}
       \includegraphics[width=0.85\textwidth]{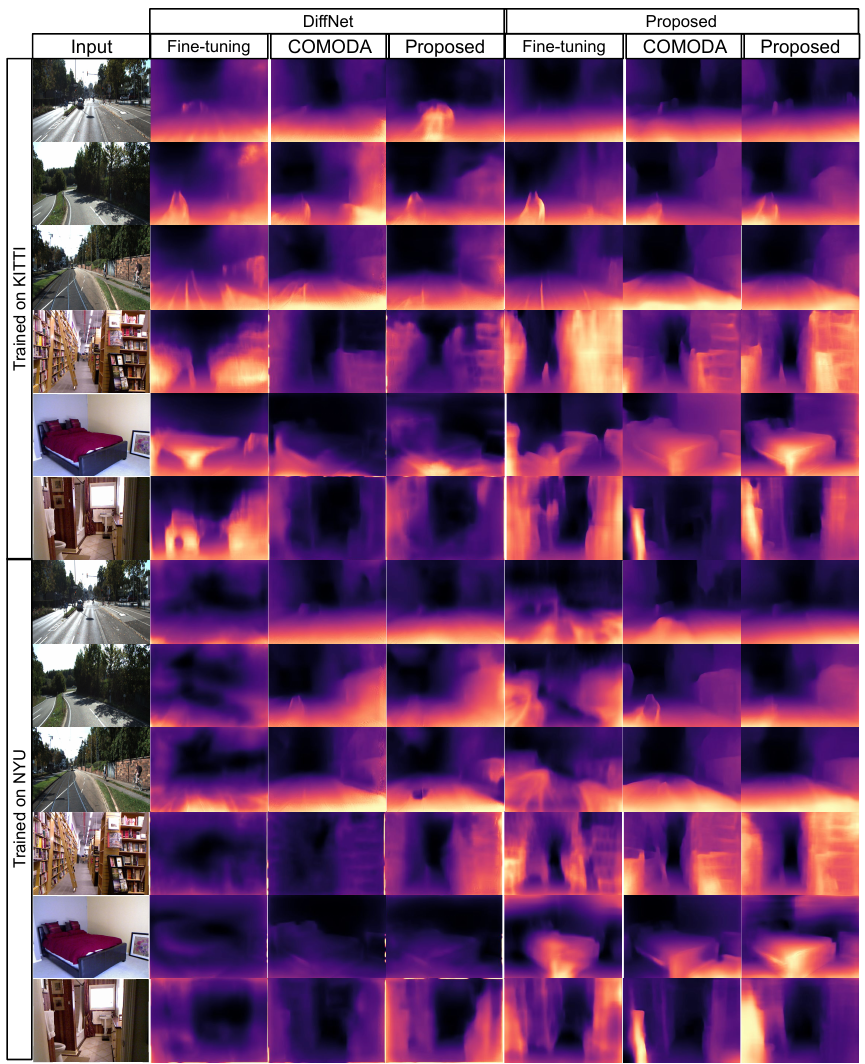}
       \caption{Qualitative results with SfM over KITTI and NYU datasets with different networks and methods for continual leanring. Best seen in color.}
       \label{fig:qual_kitti_nyu}
    \end{center}
 \end{figure*}
 \begin{figure*}[!ht]
    \begin{center}
       \includegraphics[width=0.85\textwidth]{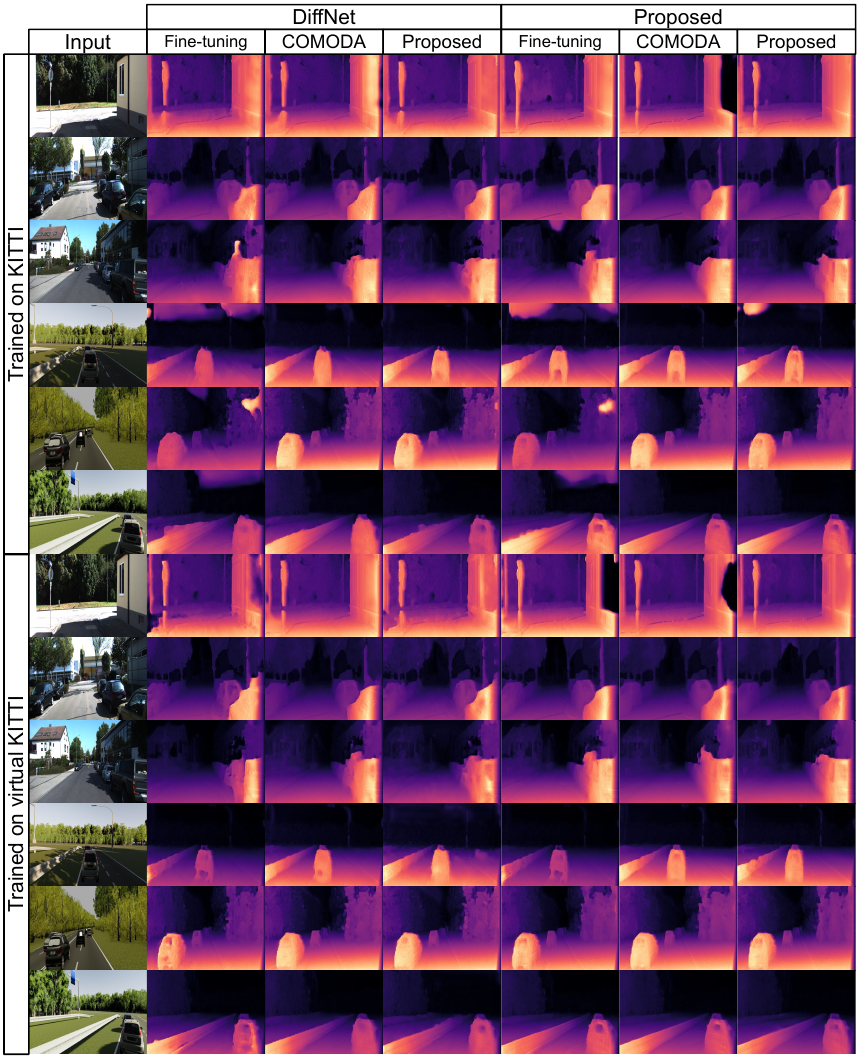}
       \caption{Qualitative results with stereo over KITTI and virtual KITTI datasets with different networks and methods for continual learning. Best seen in color.}
       \label{fig:qual_kitti_vkitti}
    \end{center}
 \end{figure*}
 
 \begin{figure*}[!ht]
 \includegraphics[width=\textwidth]{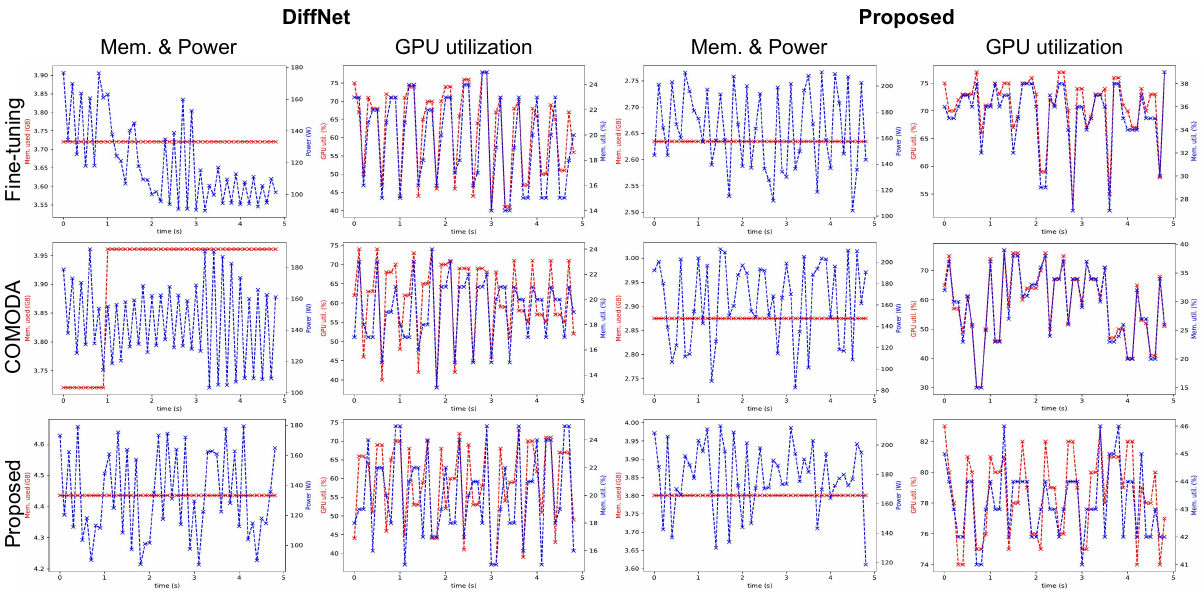}
 \centering
 \caption{Hardware performance with different networks and methods for continual learning while training SfM. Memory and GPU utilization show the percent of time over the past sample period when memory read/write was performed and a kernel was executed on the GPU, respectively.}
 \label{fig:gpu_sfm}
 \end{figure*}
 
  \begin{figure*}[!ht]
 \includegraphics[width=\textwidth]{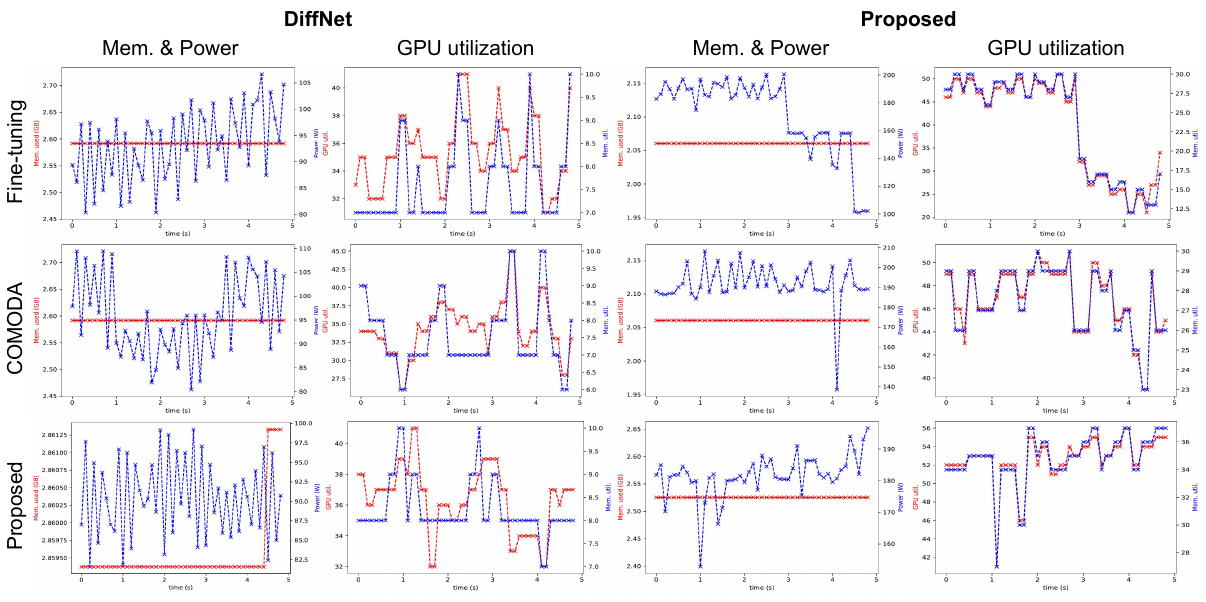}
 \centering
 \caption{Hardware performance with different networks and methods for continual learning while training stereo. Memory and GPU utilization show the percent of time over the past sample period when memory read/write was performed and a kernel was executed on the GPU, respectively.}
 \label{fig:gpu_stereo}
 \end{figure*}
 
 \subsection{Training and Evaluation Protocol}
 \noindent Online training is preceded by pre-training or warmup \cite{hayes2020remind}. Pre-training and online training data are composed of different scenes. Pre-training provides the base model for online training. For online training, data is sequentially passed without shuffling. Thus, temporal correlation exists which is more practical and challenging \cite{chrysakis2020online}. Proposed results are averaged over three runs of online training. The evaluation results are obtained by averaging the results of all frames of the test data. Further details are in the supplementary. The test data is not used for learning unlike \cite{zhang2020online} and \cite{tonioni2019learning} but for evaluation only. Thus, there is no test-time oracle. 
 
 The following metrics have been used in this work. 
  \begin{itemize} 
    \item \textbf{RMSE}: $\sqrt{\left(\frac{1}{N}\sum_x(d(x) - d^*(x))^2\right)}$
    \item \textbf{Absolute Relative}: $\frac{1}{N}\sum_x\frac{\left|d(x) - d^*(x)\right|}{d^*(x)}$
    \item $\boldsymbol{\delta(a)}$: $\frac{1}{N}\sum_x~1\left\{\max\left(\frac{d^*(x)}{d(x)}, \frac{d(x)}{d^*(x)}\right) < a\right\}$
 \end{itemize}
 where $x$, $d$ and $d^*$ represent the pixel location, obtained depth and ground-truth depth, respectively. Check supplementary for detailed results on other metrics as well. 
 
 \subsection{Setup}
 \noindent The same hyper-parameters and optimizer are used in all experiments. The Adam optimizer \cite{kingma2014adam} is used with $\beta_1 = 0.9$, $\beta_2 = 0.99$, learning rate of $10^{-4}$, batch size of 1, $\alpha_L = 0.1$, $\gamma = 10^{-2}$, $\beta_{p} = 0.15$, $\beta_{ss} = 0.85$, and $\beta_{s} = 0.1$. A single Nvidia GTX 1080Ti GPU is used and frames are resized to $320 \times 256$. 
 
 \subsection{Performance Evaluation}
 \noindent The first set of experiments are based on SfM, where pre-training is performed on roughly half of the KITTI and NYU datasets, and online training is performed on the other half of either KITTI or NYU. The results for the absolute-relative metric are shown in Table \ref{tab:sfm_stereo}. Refer to supplementary for results on other metrics. For comparison, we inlcude results of the COMODA \cite{kuznietsov2021comoda} continual learning approach, where samples from dataset used for pre-training are randomly used for replay with equal probability as the online-training samples. We apply COMODA and the proposed method of continual learning to the DiffNet \cite{zhou2021self} network and the proposed network architecture described in previous section. Online training on one dataset severely degrades performance on the other dataset if it is directly fine-tuned. However, by using the proposed approach of regularization and replay the neural network remembers information from the dataset it is not being trained on. Furthermore, plasticity is not degraded, rather, it is improved. This should come as no surprise as regularization improves forward transfer of information and replay exposes the network to difficult examples. The performance with the DiffNet architecture is generally poorer compared to the proposed architecture. It is seen that COMODA slightly outperforms the proposed method in the NDP. This shows that COMODA shows similar or better performance on the non-training dataset observed during pre-training. However, the propsed method outperforms COMODA on the TDP, which considers both the online-adaptation and the performance on previously observed scenes in the dataset used for online training. The qualitative results with online training with different continual learning strategies and network architectures are shown in Fig. \ref{fig:qual_kitti_nyu}. The results clearly demonstrate that the fine-tuning does degrade the output drastically. Furthermore, the proposed method of continual learning with the proposed architecture generally produces superior results compared to other alternatives.
 
 Note that these results should not be compared with methods that follow the paradigm of training followed by testing. With KITTI in conventional setting, the network is trained over the KITTI training data for multiple epochs and then evaluated over the test data. In the current scenario, the network is pre-trained over half of KITTI and NYU datasets for multiple epochs, and then trained over the other half of the KITTI dataset for one epoch only during online training. 
 
 For evaluation over stereo, pre-training is performed over half of the KITTI and virtual KITTI datasets, and online training on the other half of either KITTI or virtual KITTI datasets. The results are shown in Table \ref{tab:sfm_stereo}. Again we see a significant improvement is seen in performance with the proposed method. Qualitative results are shown in Fig. \ref{fig:qual_kitti_vkitti}. We note that the proposed method does outperform competing methods, however, the performance of competing methods is not as poor as in the case of SfM. This is an interesting result. We believe that this is because in case of SfM the networks are trained to estimate disparity from a single image. Thus, the network is required to memorize the size-to-depth relationship of different objects in the scene. Once the scene changes, new objects appear which have a very different size-to-depth relationship compared to previously observed objects, resulting in the networks perform poorly. Stereo-based approach, however, learns to estimate disparity from left and right images, and only uses size-to-depth relationship where disparity estimation is not possible, for example, textureless or occluded regions. The disparity estimation process remains the same across scenes, therefore, stereo-based depth estimation is not as severely affected by observing new scenes as SfM.
 
 To further analyze the performance of these methods, we show the results of different methods and networks on individual scenes from the online training datasets after online training in Fig. \ref{fig:ind_scenes}. We see that the proposed network architecture trained with the proposed method for continual learning shows better performance on individual scenes observed sequentially during training with both SfM and stereo. This is because the proposed approach not only remembers scenes from pre-training but also has mechanisms to remember older scenes observed during onling training.
 
 We conduct another experiment to validate the performance of the proposed method. In \cite{zhang2020online}, the authors use SfM where they first pretrain the model on the KITTI dataset over multiple epochs and then online train over the virtual KITTI dataset for a single epoch. They also perform test-time adaptation, i.e., they use the sequences used for testing to learn while testing. We also test our method under their experimental settings and the results are shown in Table \ref{tab:sfm_kitti_vkitti}. However, test-time adaptation is not performed with the proposed method. Using the complete video sequence in testing provides useful context, which maybe taken as test-time oracle. Unlike \cite{zhang2020online}, results where the challenging \texttt{fog} and \texttt{rain} sequences of the virtual KITTI dataset are included in the online training are also included. The results show that the proposed method outperforms competing methods and provides a larger percentage increase compared to fine-tuning.
 
 \subsection{GPU Performance}
 \noindent One important aspect worth attention is the overhead incurred by the proposed method compared to fine tuning. To analyze this, we trace the GPU performance with different metrics for 5 seconds while online training. The results for SfM and stereo are shown in Fig. \ref{fig:gpu_sfm} and \ref{fig:gpu_stereo}, respectively. All the methods for continual learning with the proposed network architecture provide 12fps and 17fps for SfM and stereo, respectively, which is higher than the frame rate of 10 of the KITTI dataset. The fps goes down to 4fps and 5fps with DiffNet \cite{zhou2021self} in the same order. This shows that although there is still room for improvement, there is no overhead in terms of speed with the proposed method for continual learning. The figures show that there is only a slight increase in the GPU memory consumption with the proposed approach for continual learning. The maximum GPU memory required with the proposed method for continual learning and the proposed network architecture is 3.8GB, which can be easily accommodated on low-end GPUs for deep learning such as NVIDIA GeForce GTX 1650 Ti Mobile. For all other metrics, the proposed method for continual learning of depth has the same resource utilization as not using any method for continual learning at all. The power consumption with the proposed network architecture is higher compared to the power consumption with DiffNet. This is because the proposed network architecture operates at a higher frame rate compared to DiffNet.  
 
 \section{Conclusion}
 \noindent A method is presented for learning depth of a scene while deployed and not forgetting about the past. Novel regularization and replay methods are proposed to this end. The method learns depth from a stream of monocular or stereo video-frames, and avoids catastrophic forgetting. Experiments show that the proposed method has superior performance with both SfM and stereo. This work is a step towards autonomous scene-perception without any supervision. 
 
 \appendix 
 Based on (1), we can write
 
 \begin{equation} 
 P(T(t) = T(t - 1)) = \mathcal{N}(L_t; \mu_{t-1}, \sigma_{t-1})
 \end{equation}
 
 or 
 
 \begin{equation} 
 P(T(t) = T(t - 1)) = \frac{1}{\sqrt{2\pi}\sigma_{t-1}}e^{\frac{-(L_t - \mu_{t-1})^2}{2\sigma_{t-1}^2}}.
 \end{equation}
 
 Let 
 \begin{equation}
     B = \frac{-(L_t - \mu_{t-1})^2}{2\sigma_{t-1}^2}
 \end{equation}
 
 \noindent and 
 
 \begin{equation}
     C = \sqrt{2\pi}\sigma_{t-1}.
 \end{equation}
 
 \noindent Thus 
 \begin{equation} 
    P(T(t) = T(t-1)) = \frac{exp(B)}{C}.
\end{equation}

\noindent Taking logarithm of both sides 

\begin{equation}
    \log(P(T(t) = T(t - 1)) = \log\left(\frac{exp(B)}{C}\right).
\end{equation}

\noindent Using the properties of $\log$, and substituting $A$ and $B$, we can write  
\begin{equation}
    -\log(P(T(t) = T(t - 1)) = \frac{1}{2}\frac{(L_t - \mu_{t-1})^2}{\sigma_{t-1}^2} + \log(\sqrt{2\pi}\sigma_{t-1}). 
\end{equation}

\bibliographystyle{IEEEtran}
 \bibliography{main}

\newpage
\section{Datasets and Scenes}
 The datasets are divided into two parts: Pre-training and Online training. Scenes are assigned to pre-training and online training such that the number of scenes remains roughly the same in both.
 \subsection{SfM Experiments}
 \subsubsection{KITTI Dataset}
 The KITTI dataset is already divided into multiple categories based on the location of the vehicle. Majority of these scenes are from the \texttt{residential} category. It is very difficult to have a reasonable division of the KITTI dataset based on categories. Thus, residential sequences were divided into two parts: \texttt{residential\_1} and \texttt{residential\_2}.
 \begin{itemize} 
    \item Training frames: 44764
    \item Testing frames: 697
    \item Pre-training 
    \begin{itemize} 
       \item Scenes: Road, residential\_1
       \item Frames: 24920 
    \end{itemize} 
    \item Online Training
    \begin{itemize} 
       \item Scenes: Residential\_2, city, campus 
       \item Frames: 19844
    \end{itemize} 
 \end{itemize}
 
 \subsubsection{NYU Dataset}
 \begin{itemize} 
    \item Training frames: 69383
    \item Testing frames: 654
    \item Pre-training 
    \begin{itemize} 
       \item Scenes: From \emph{basement} to \emph{indoor\_balcony}
       \item Frames: 29596 
    \end{itemize} 
    \item Online Training
    \begin{itemize} 
       \item Scenes: The rest  
       \item Frames: 39787
    \end{itemize} 
 \end{itemize}
 
 \subsection{Stereo Experiments}
 \subsubsection{KITTI Dataset}
 \begin{itemize} 
    \item Training frames: 41888
    \item Testing frames: 697
    \item Pre-training 
    \begin{itemize} 
       \item Scenes: Road, residential\_1
       \item Frames: 21696 
    \end{itemize} 
    \item Online Training
    \begin{itemize} 
       \item Scenes: Residential\_2, city, campus 
       \item Frames: 20192
    \end{itemize} 
 \end{itemize}
 
 \subsubsection{vKITTI Dataset}
 \begin{itemize} 
    \item Training frames: 21620
    \item Testing frames: 2100
    \item Pre-training 
    \begin{itemize} 
       \item Scenes: 15-deg-left, 15-deg-right, clone, fog, morning 
       \item Frames: 9580 
    \end{itemize} 
    \item Online Training
    \begin{itemize} 
       \item Scenes: 30-deg-left, 30-deg-right, overcast, rain, sunset  
       \item Frames: 9580
    \end{itemize} 
 \end{itemize}
 
 \subsection{Test Frames Categories}
 For vKITTI, the directory from which the frame is taken gives us the test frame category. The text file \href{https://github.com/nianticlabs/monodepth2/blob/master/splits/eigen/test\_files.txt}{here} was used for first finding the sequence to which the frame belongs, and then find the category to which the sequence belongs using the KITTI website. For NYU, the .mat file provided \href{http://horatio.cs.nyu.edu/mit/silberman/nyu\_depth\_v2/nyu\_depth\_v2\_labeled.mat}{here} was used to find the categories of the test frames.
 
 \section{Evaluation Metrics}
 Let $N$ be the total number of pixels in a frame used for evaluation and $x$ be the pixel index. Let $d$ and $d^*$ represent the obtained and the ground truth distance, respectively. Also, $1\{.\}$ is the indicator function, which returns $1$ if the condition in the parentheses is true, otherwise returns $0$.
 
 \begin{itemize} 
    \item \textbf{RMSE}: $\sqrt{\left(\frac{1}{N}\sum_x(d(x) - d^*(x))^2\right)}$
    \item \textbf{Absolute Relative}: $\frac{1}{N}\sum_x\frac{\left|d(x) - d^*(x)\right|}{d^*(x)}$
    \item \textbf{Square Relative}: $\frac{1}{N}\sum_x\frac{(d(x) - d^*(x))^2}{d^*(x)}$
    \item \textbf{Log RMSE}: $\frac{1}{N}\sum_x(\log(d(x)) - \log(d^*(x)))$
    \item $\boldsymbol{\delta(a)}$: $\frac{1}{N}\sum_x~1\left\{\max\left(\frac{d^*(x)}{d(x)}, \frac{d(x)}{d^*(x)}\right) < a\right\}$
 \end{itemize}
 
 \section{Evaluation Protocol}
 The results in the manuscript are given for the whole dataset as well as domains within the dataset. These are described as follows. 
 
 \subsection{Training Dataset Performance (TDP) Results}
 The results on the test data of the dataset over which training is performed. For example, if training is performed on KITTI then TDP results show the results over the test data of KITTI dataset after online training is complete. 
 
 \subsection{Non-training Dataset Performanc (NDP) Results}
 The results on the test data of the dataset other than the one over which training is performed. For example, if training is performed on KITTI then NDP results show the results over the test data of NYU dataset after online training is complete.
 
 \section{Storage Requirements}
 It is noted that $320 \times 256$ \texttt{.jpg} images are approximately 20KB in size. This is the resolution used in our experiments. The number of frames stored for replay did not exceed 1500 in any of our experiments.
 \begin{itemize}
 \item KITTI + NYU SfM Experiments: The storage space required to maintain the pre-training dataset is approximately 1.04GB. An additional 12086 frames can be maintained for replay. 
 
 \item KITTI + vKITTI SfM Experiments: The storage space required to maintain the pre-training dataset (KITTI) is approximately 0.85GB. An additional 16939 frames can be maintained for replay. 
 
 \item KITTI + vKITTI Stereo Experiments: The storage space required to maintain the stereo pre-training dataset is approximately 1.19GB. An additional 8045 stereo images can be maintained for replay.
 \end{itemize}

 \section{Ablative Results}
 The effect of regularization and replay is shown in Table \ref{tab:ablative}. The table shows that the combination of replay and regularization provide the overall best results.

 \begin{table}
    \begin{center}
       \small
       \begin{tabular}{llcc}
          \hline 
          Meth. & Data & TDP & NDP\\ 
          \hline \hline 
          FT & KITTI & 7.3054 & 11805 \\
          Reg. & KITTI & 6.6603 & 0.6955 \\
          Rep. & KITTI & 5.9583 & 0.6782 \\
          Prop. & KITTI & \textbf{5.8085} & \textbf{0.6603} \\
          \hline
          FT & NYU & 0.7688 & 10.2060 \\
          Reg. & NYU & 0.7010 & 8.5240 \\
          Rep. & NYU & 0.6368 & 7.3262 \\
          Prop. & NYU & \textbf{0.6270} & \textbf{5.9958} \\
          \hline
       \end{tabular}
    \end{center}
    \caption{RMSE under with different components of the proposed approach for a single run. FT, Reg., Rep. and Prop. stand for fine tuning, regularization, replay and proposed method, respectively. KITTI + NYU was used for pre-training.}
    \label{tab:ablative}
 \end{table}
 
 \section{Additional Results}
 \begin{table*}
    \caption{Absolute Relative of SfM-based depth estimation. \textit{FT} stands for fine tuning.}
    \label{tab:kitti_nyu}
    \begin{center}
       \small
       \begin{tabular}{lllcc}
          \hline 
          \begin{tabular}[x]{@{}c@{}}Online\\Train\end{tabular} 
           & Method & Network & TDP & NDP\\ 
          \hline \hline
KITTI & FT & DiffNet \cite{zhou2021self} &0.2214 $\pm$ 0.0000 & 0.5746 $\pm$ 0.0000 \\
KITTI & Prop. & DiffNet \cite{zhou2021self} &0.1916 $\pm$ 0.0024 & 0.2694 $\pm$ 0.0213 \\
KITTI & \cite{kuznietsov2021comoda} & DiffNet \cite{zhou2021self} &0.1904 $\pm$ 0.0140 & 0.2265 $\pm$ 0.0056 \\
KITTI & FT & Prop. &0.1895 $\pm$ 0.0000 & 0.3504 $\pm$ 0.0000 \\
KITTI & Prop. & Prop. &0.1543 $\pm$ 0.0034 & 0.1952 $\pm$ 0.0021 \\
KITTI & \cite{kuznietsov2021comoda} & Prop. &0.1580 $\pm$ 0.0064 & 0.1911 $\pm$ 0.0010 \\
\hline 
NYU & FT & DiffNet \cite{zhou2021self} &0.3243 $\pm$ 0.0000 & 0.3088 $\pm$ 0.0000 \\
NYU & Prop. & DiffNet \cite{zhou2021self} &0.2379 $\pm$ 0.0030 & 0.1995 $\pm$ 0.0176 \\
NYU & \cite{kuznietsov2021comoda} & DiffNet \cite{zhou2021self} &0.2243 $\pm$ 0.0016 & 0.1695 $\pm$ 0.0013 \\ 
NYU & FT & Prop. &0.2430 $\pm$ 0.0000 & 0.3336 $\pm$ 0.0000 \\
NYU & Prop. & Prop. &0.1872 $\pm$ 0.0058 & 0.1624 $\pm$ 0.0044 \\
NYU & \cite{kuznietsov2021comoda} & Prop. &0.1912 $\pm$ 0.0003 & 0.1586 $\pm$ 0.0037 \\
          \hline
       \end{tabular}
    \end{center}
 \end{table*}
 
 
  \begin{table*}
    \caption{RMSE of SfM-based depth estimation. \textit{FT} stands for fine tuning.}
    \label{tab:kitti_nyu}
    \begin{center}
       \small
       \begin{tabular}{lllcc}
          \hline 
          \begin{tabular}[x]{@{}c@{}}Online\\Train\end{tabular} 
           & Method & Network & TDP & NDP\\ 
          \hline \hline
KITTI & FT & DiffNet \cite{zhou2021self} &7.9542 $\pm$ 0.0000 & 2.2098 $\pm$ 0.0000 \\
KITTI & Prop. & DiffNet \cite{zhou2021self} &6.3588 $\pm$ 0.1508 & 0.8783 $\pm$ 0.0733 \\
KITTI & \cite{kuznietsov2021comoda} & DiffNet \cite{zhou2021self} &6.4767 $\pm$ 0.5267 & 0.7731 $\pm$ 0.0125 \\
KITTI & FT & Prop. &7.3054 $\pm$ 0.0000 & 1.1805 $\pm$ 0.0000 \\
KITTI & Prop. & Prop. &5.8085 $\pm$ 0.1168 & 0.6603 $\pm$ 0.0015 \\
KITTI & \cite{kuznietsov2021comoda} & Prop. &6.2494 $\pm$ 0.1864 & 0.6629 $\pm$ 0.0057 \\
\hline 
NYU & FT & DiffNet \cite{zhou2021self} &1.0214 $\pm$ 0.0000 & 10.7680 $\pm$ 0.0000 \\
NYU & Prop. & DiffNet \cite{zhou2021self} &0.7942 $\pm$ 0.0051 & 6.6515 $\pm$ 0.4935 \\
NYU & \cite{kuznietsov2021comoda} & DiffNet \cite{zhou2021self} &0.7660 $\pm$ 0.0115 & 6.3401 $\pm$ 0.3011 \\
NYU & FT & Prop. &0.7688 $\pm$ 0.0000 & 10.2060 $\pm$ 0.0000 \\
NYU & Prop. & Prop. &0.6270 $\pm$ 0.0115 & 5.9958 $\pm$ 0.0847 \\
NYU & \cite{kuznietsov2021comoda} & Prop. &0.6727 $\pm$ 0.0025 & 6.2292 $\pm$ 0.1704 \\
          \hline
       \end{tabular}
    \end{center}
 \end{table*}
 
  
  \begin{table*}
    \caption{Square Relative of SfM-based depth estimation. \textit{FT} stands for fine tuning.}
    \label{tab:kitti_nyu}
    \begin{center}
       \small
       \begin{tabular}{lllcc}
          \hline 
          \begin{tabular}[x]{@{}c@{}}Online\\Train\end{tabular} 
           & Method & Network & TDP & NDP\\ 
          \hline \hline
KITTI & FT & DiffNet \cite{zhou2021self} &1.9197 $\pm$ 0.0000 & 1.9714 $\pm$ 0.0000 \\
KITTI & Prop. & DiffNet \cite{zhou2021self} &1.4904 $\pm$ 0.0390 & 0.3246 $\pm$ 0.0713 \\
KITTI & \cite{kuznietsov2021comoda} & DiffNet \cite{zhou2021self} &1.4904 $\pm$ 0.1844 & 0.2315 $\pm$ 0.0122 \\
KITTI & FT & Prop. &1.5520 $\pm$ 0.0000 & 0.5699 $\pm$ 0.0000 \\
KITTI & Prop. & Prop. &1.1420 $\pm$ 0.0419 & 0.1748 $\pm$ 0.0029 \\
KITTI & \cite{kuznietsov2021comoda} & Prop. &1.2297 $\pm$ 0.0744 & 0.1745 $\pm$ 0.0027 \\
\hline 
NYU & FT & DiffNet \cite{zhou2021self} &0.4248 $\pm$ 0.0000 & 3.2486 $\pm$ 0.0000 \\
NYU & Prop. & DiffNet \cite{zhou2021self} &0.2490 $\pm$ 0.0052 & 1.5412 $\pm$ 0.1497 \\
NYU & \cite{kuznietsov2021comoda} & DiffNet \cite{zhou2021self} &0.2295 $\pm$ 0.0046 & 1.3228 $\pm$ 0.0281 \\
NYU & FT & Prop. &0.2432 $\pm$ 0.0000 & 3.1929 $\pm$ 0.0000 \\
NYU & Prop. & Prop. &0.1624 $\pm$ 0.0095 & 1.2044 $\pm$ 0.0390 \\
NYU & \cite{kuznietsov2021comoda} & Prop. &0.1787 $\pm$ 0.0010 & 1.2233 $\pm$ 0.0544 \\
          \hline
       \end{tabular}
    \end{center}
 \end{table*}
 
 
  
  \begin{table*}
    \caption{Log RMSE of SfM-based depth estimation. \textit{FT} stands for fine tuning.}
    \label{tab:kitti_nyu}
    \begin{center}
       \small
       \begin{tabular}{lllcc}
          \hline 
          \begin{tabular}[x]{@{}c@{}}Online\\Train\end{tabular} 
           & Method & Network & TDP & NDP\\ 
          \hline \hline
KITTI & FT & DiffNet \cite{zhou2021self} &0.3276 $\pm$ 0.0000 & 0.5613 $\pm$ 0.0000 \\
KITTI & Prop. & DiffNet \cite{zhou2021self} &0.2750 $\pm$ 0.0055 & 0.3290 $\pm$ 0.0149 \\
KITTI & \cite{kuznietsov2021comoda} & DiffNet \cite{zhou2021self} &0.2779 $\pm$ 0.0243 & 0.3061 $\pm$ 0.0026 \\
KITTI & FT & Prop. &0.2927 $\pm$ 0.0000 & 0.3921 $\pm$ 0.0000 \\
KITTI & Prop. & Prop. &0.2351 $\pm$ 0.0006 & 0.2401 $\pm$ 0.0016 \\
KITTI & \cite{kuznietsov2021comoda} & Prop. &0.2477 $\pm$ 0.0084 & 0.2396 $\pm$ 0.0017 \\
\hline 
NYU & FT & DiffNet \cite{zhou2021self} &0.3862 $\pm$ 0.0000 & 0.4634 $\pm$ 0.0000 \\
NYU & Prop. & DiffNet \cite{zhou2021self} &0.3088 $\pm$ 0.0029 & 0.2804 $\pm$ 0.0214 \\
NYU & \cite{kuznietsov2021comoda} & DiffNet \cite{zhou2021self} &0.2955 $\pm$ 0.0062 & 0.2567 $\pm$ 0.0073 \\
NYU & FT & Prop. &0.2883 $\pm$ 0.0000 & 0.4518 $\pm$ 0.0000 \\
NYU & Prop. & Prop. &0.2296 $\pm$ 0.0050 & 0.2398 $\pm$ 0.0045 \\
NYU & \cite{kuznietsov2021comoda} & Prop. &0.2406 $\pm$ 0.0007 & 0.2467 $\pm$ 0.0048 \\
          \hline
       \end{tabular}
    \end{center}
 \end{table*}
 

  \begin{table*}
    \caption{$\delta(1.25)$ of SfM-based depth estimation. \textit{FT} stands for fine tuning.}
    \label{tab:kitti_nyu}
    \begin{center}
       \small
       \begin{tabular}{lllcc}
          \hline 
          \begin{tabular}[x]{@{}c@{}}Online\\Train\end{tabular} 
           & Method & Network & TDP & NDP\\ 
          \hline \hline
KITTI & FT & DiffNet \cite{zhou2021self} &0.6475 $\pm$ 0.0000 & 0.2894 $\pm$ 0.0000 \\
KITTI & Prop. & DiffNet \cite{zhou2021self} &0.7078 $\pm$ 0.0040 & 0.5688 $\pm$ 0.0299 \\
KITTI & \cite{kuznietsov2021comoda} & DiffNet \cite{zhou2021self} &0.7037 $\pm$ 0.0377 & 0.6373 $\pm$ 0.0064 \\
KITTI & FT & Prop. &0.7110 $\pm$ 0.0000 & 0.4462 $\pm$ 0.0000 \\
KITTI & Prop. & Prop. &0.7835 $\pm$ 0.0027 & 0.7029 $\pm$ 0.0041 \\
KITTI & \cite{kuznietsov2021comoda} & Prop. &0.7691 $\pm$ 0.0157 & 0.7047 $\pm$ 0.0066 \\
\hline 
NYU & FT & DiffNet \cite{zhou2021self} &0.4746 $\pm$ 0.0000 & 0.4509 $\pm$ 0.0000 \\
NYU & Prop. & DiffNet \cite{zhou2021self} &0.6199 $\pm$ 0.0081 & 0.6835 $\pm$ 0.0458 \\
NYU & \cite{kuznietsov2021comoda} & DiffNet \cite{zhou2021self} &0.6446 $\pm$ 0.0034 & 0.7446 $\pm$ 0.0105 \\
NYU & FT & Prop. &0.6063 $\pm$ 0.0000 & 0.4050 $\pm$ 0.0000 \\
NYU & Prop. & Prop. &0.7284 $\pm$ 0.0078 & 0.7671 $\pm$ 0.0069 \\
NYU & \cite{kuznietsov2021comoda} & Prop. &0.7055 $\pm$ 0.0011 & 0.7688 $\pm$ 0.0085 \\
          \hline
       \end{tabular}
    \end{center}
 \end{table*}
 

  \begin{table*}
    \caption{$\delta(1.25^2)$ of SfM-based depth estimation. \textit{FT} stands for fine tuning.}
    \label{tab:kitti_nyu}
    \begin{center}
       \small
       \begin{tabular}{lllcc}
          \hline 
          \begin{tabular}[x]{@{}c@{}}Online\\Train\end{tabular} 
           & Method & Network & TDP & NDP\\ 
          \hline \hline
KITTI & FT & DiffNet \cite{zhou2021self} &0.8489 $\pm$ 0.0000 & 0.5540 $\pm$ 0.0000 \\
KITTI & Prop. & DiffNet \cite{zhou2021self} &0.8930 $\pm$ 0.0050 & 0.8328 $\pm$ 0.0224 \\
KITTI & \cite{kuznietsov2021comoda} & DiffNet \cite{zhou2021self} &0.8956 $\pm$ 0.0226 & 0.8638 $\pm$ 0.0039 \\
KITTI & FT & Prop. &0.8807 $\pm$ 0.0000 & 0.7442 $\pm$ 0.0000 \\
KITTI & Prop. & Prop. &0.9278 $\pm$ 0.0011 & 0.9140 $\pm$ 0.0029 \\
KITTI & \cite{kuznietsov2021comoda} & Prop. &0.9180 $\pm$ 0.0073 & 0.9130 $\pm$ 0.0014 \\
\hline 
NYU & FT & DiffNet \cite{zhou2021self} &0.7662 $\pm$ 0.0000 & 0.7384 $\pm$ 0.0000 \\
NYU & Prop. & DiffNet \cite{zhou2021self} &0.8572 $\pm$ 0.0015 & 0.8892 $\pm$ 0.0213 \\
NYU & \cite{kuznietsov2021comoda} & DiffNet \cite{zhou2021self} &0.8671 $\pm$ 0.0037 & 0.9098 $\pm$ 0.0070 \\
NYU & FT & Prop. &0.8696 $\pm$ 0.0000 & 0.7034 $\pm$ 0.0000 \\
NYU & Prop. & Prop. &0.9220 $\pm$ 0.0050 & 0.9243 $\pm$ 0.0020 \\
NYU & \cite{kuznietsov2021comoda} & Prop. &0.9127 $\pm$ 0.0009 & 0.9178 $\pm$ 0.0047 \\

          \hline
       \end{tabular}
    \end{center}
 \end{table*}

  \begin{table*}
    \caption{Absolute Relative of Stereo-based depth estimation. \textit{FT} stands for fine tuning.}
    \label{tab:kitti_nyu}
    \begin{center}
       \small
       \begin{tabular}{lllcc}
          \hline 
          \begin{tabular}[x]{@{}c@{}}Online\\Train\end{tabular} 
           & Method & Network & TDP & NDP\\ 
          \hline \hline
          KITTI & FT & DiffNet \cite{zhou2021self} & 0.1932 $\pm$ 0.0000 & 0.1996 $\pm$ 0.0000 \\
KITTI & Prop. & DiffNet \cite{zhou2021self} & 0.2123 $\pm$ 0.0192 & 0.1902 $\pm$ 0.0113 \\
KITTI & \cite{kuznietsov2021comoda} & DiffNet \cite{zhou2021self} & 0.2098 $\pm$ 0.0433 & 0.1656 $\pm$ 0.0067 \\
KITTI & FT & Prop. & 0.1920 $\pm$ 0.0000 & 0.1980 $\pm$ 0.0000 \\
KITTI & Prop. & Prop. & 0.1825 $\pm$ 0.0037 & 0.1660 $\pm$ 0.0037 \\
KITTI & \cite{kuznietsov2021comoda} & Prop. & 0.1962 $\pm$ 0.0220 & 0.1685 $\pm$ 0.0050 \\
\hline 
vKITTI & FT & DiffNet \cite{zhou2021self} & 0.1860 $\pm$ 0.0000 & 0.2401 $\pm$ 0.0000 \\
vKITTI & Prop. & DiffNet \cite{zhou2021self} & 0.1765 $\pm$ 0.0025 & 0.2190 $\pm$ 0.0200 \\
vKITTI & \cite{kuznietsov2021comoda} & DiffNet \cite{zhou2021self} & 0.1803 $\pm$ 0.0067 & 1.4442 $\pm$ 1.7665 \\
vKITTI & FT & Prop. & 0.1991 $\pm$ 0.0000 & 0.2090 $\pm$ 0.0000 \\
vKITTI & Prop. & Prop. & 0.1653 $\pm$ 0.0038 & 0.1770 $\pm$ 0.0066 \\
vKITTI & \cite{kuznietsov2021comoda} & Prop. & 0.1834 $\pm$ 0.0174 & 0.3177 $\pm$ 0.1860 \\
          \hline
       \end{tabular}
    \end{center}
 \end{table*}

  \begin{table*}
    \caption{RMSE of Stereo-based depth estimation. \textit{FT} stands for fine tuning.}
    \label{tab:kitti_nyu}
    \begin{center}
       \small
       \begin{tabular}{lllcc}
          \hline 
          \begin{tabular}[x]{@{}c@{}}Online\\Train\end{tabular} 
           & Method & Network & TDP & NDP\\ 
          \hline \hline
          KITTI & FT & DiffNet \cite{zhou2021self} & 7.1615 $\pm$ 0.0000 & 9.2327 $\pm$ 0.0000 \\
KITTI & Prop. & DiffNet \cite{zhou2021self} & 8.3708 $\pm$ 2.5778 & 8.0995 $\pm$ 0.2314 \\
KITTI & \cite{kuznietsov2021comoda} & DiffNet \cite{zhou2021self} & 10.4766 $\pm$ 5.8117 & 7.4932 $\pm$ 0.0850 \\
KITTI & FT & Prop. & 7.5129 $\pm$ 0.0000 & 8.8856 $\pm$ 0.0000 \\
KITTI & Prop. & Prop. & 6.3309 $\pm$ 0.2618 & 7.4888 $\pm$ 0.2729 \\
KITTI & \cite{kuznietsov2021comoda} & Prop. & 7.9609 $\pm$ 1.7703 & 7.4235 $\pm$ 0.0857 \\
\hline 
vKITTI & FT & DiffNet \cite{zhou2021self} & 10.8090 $\pm$ 0.0000 & 11.2852 $\pm$ 0.0000 \\
vKITTI & Prop. & DiffNet \cite{zhou2021self} & 8.7305 $\pm$ 0.7399 & 7.9432 $\pm$ 1.6038 \\
vKITTI & \cite{kuznietsov2021comoda} & DiffNet \cite{zhou2021self} & 8.1609 $\pm$ 0.4309 & 64.6970 $\pm$ 80.7587 \\
vKITTI & FT & Prop. & 15.5476 $\pm$ 0.0000 & 11.3597 $\pm$ 0.0000 \\
vKITTI & Prop. & Prop. & 7.5368 $\pm$ 0.1035 & 6.6097 $\pm$ 0.4084 \\
vKITTI & \cite{kuznietsov2021comoda} & Prop. & 14.6367 $\pm$ 8.4462 & 26.1553 $\pm$ 26.3570 \\
          \hline
       \end{tabular}
    \end{center}
 \end{table*}

  \begin{table*}
    \caption{Square Relative of Stereo-based depth estimation. \textit{FT} stands for fine tuning.}
    \label{tab:kitti_nyu}
    \begin{center}
       \small
       \begin{tabular}{lllcc}
          \hline 
          \begin{tabular}[x]{@{}c@{}}Online\\Train\end{tabular} 
           & Method & Network & TDP & NDP\\ 
          \hline \hline
KITTI & FT & DiffNet \cite{zhou2021self} & 2.1568 $\pm$ 0.0000 & 2.5450 $\pm$ 0.0000 \\
KITTI & Prop. & DiffNet \cite{zhou2021self} & 58.9297 $\pm$ 80.4640 & 2.3211 $\pm$ 0.4232 \\
KITTI & \cite{kuznietsov2021comoda} & DiffNet \cite{zhou2021self} & 3934.6596 $\pm$ 5560.9138 & 1.7403 $\pm$ 0.0675 \\
KITTI & FT & Prop. & 2.4475 $\pm$ 0.0000 & 2.3190 $\pm$ 0.0000 \\
KITTI & Prop. & Prop. & 2.0111 $\pm$ 0.4810 & 1.7943 $\pm$ 0.0843 \\
KITTI & \cite{kuznietsov2021comoda} & Prop. & 623.9802 $\pm$ 876.0476 & 1.7083 $\pm$ 0.0845 \\
\hline 
vKITTI & FT & DiffNet \cite{zhou2021self} & 4.0448 $\pm$ 0.0000 & 10.7284 $\pm$ 0.0000 \\
vKITTI & Prop. & DiffNet \cite{zhou2021self} & 2.6488 $\pm$ 0.9602 & 24.4938 $\pm$ 30.5706 \\
vKITTI & \cite{kuznietsov2021comoda} & DiffNet \cite{zhou2021self} & 5.7252 $\pm$ 4.7819 & 280307.3590 $\pm$ 396372.1155 \\
vKITTI & FT & Prop. & 17.3129 $\pm$ 0.0000 & 18.1648 $\pm$ 0.0000 \\
vKITTI & Prop. & Prop. & 1.9370 $\pm$ 0.1294 & 2.1477 $\pm$ 0.3456 \\
vKITTI & \cite{kuznietsov2021comoda} & Prop. & 106.3278 $\pm$ 144.5834 & 112334.4625 $\pm$ 158840.1505 \\
          \hline
       \end{tabular}
    \end{center}
 \end{table*}

  \begin{table*}
    \caption{Log RMSE of Stereo-based depth estimation. \textit{FT} stands for fine tuning.}
    \label{tab:kitti_nyu}
    \begin{center}
       \small
       \begin{tabular}{lllcc}
          \hline 
          \begin{tabular}[x]{@{}c@{}}Online\\Train\end{tabular} 
           & Method & Network & TDP & NDP\\ 
          \hline \hline
KITTI & FT & DiffNet \cite{zhou2021self} & 0.2992 $\pm$ 0.0000 & 0.3515 $\pm$ 0.0000 \\
KITTI & Prop. & DiffNet \cite{zhou2021self} & 0.2999 $\pm$ 0.0033 & 0.3223 $\pm$ 0.0119 \\
KITTI & \cite{kuznietsov2021comoda} & DiffNet \cite{zhou2021self} & 0.2855 $\pm$ 0.0075 & 0.2990 $\pm$ 0.0050 \\
KITTI & FT & Prop. & 0.3004 $\pm$ 0.0000 & 0.3488 $\pm$ 0.0000 \\
KITTI & Prop. & Prop. & 0.2853 $\pm$ 0.0023 & 0.2984 $\pm$ 0.0083 \\
KITTI & \cite{kuznietsov2021comoda} & Prop. & 0.2840 $\pm$ 0.0030 & 0.2966 $\pm$ 0.0048 \\
\hline 
vKITTI & FT & DiffNet \cite{zhou2021self} & 0.3086 $\pm$ 0.0000 & 0.3307 $\pm$ 0.0000 \\
vKITTI & Prop. & DiffNet \cite{zhou2021self} & 0.3152 $\pm$ 0.0102 & 0.3070 $\pm$ 0.0086 \\
vKITTI & \cite{kuznietsov2021comoda} & DiffNet \cite{zhou2021self} & 0.3040 $\pm$ 0.0051 & 0.3156 $\pm$ 0.0328 \\
vKITTI & FT & Prop. & 0.3220 $\pm$ 0.0000 & 0.3043 $\pm$ 0.0000 \\
vKITTI & Prop. & Prop. & 0.2968 $\pm$ 0.0053 & 0.2816 $\pm$ 0.0064 \\
vKITTI & \cite{kuznietsov2021comoda} & Prop. & 0.2953 $\pm$ 0.0037 & 0.2853 $\pm$ 0.0014 \\
          \hline
       \end{tabular}
    \end{center}
 \end{table*}

  \begin{table*}
    \caption{$\delta(1.25)$ of Stereo-based depth estimation. \textit{FT} stands for fine tuning.}
    \label{tab:kitti_nyu}
    \begin{center}
       \small
       \begin{tabular}{lllcc}
          \hline 
          \begin{tabular}[x]{@{}c@{}}Online\\Train\end{tabular} 
           & Method & Network & TDP & NDP\\ 
          \hline \hline
KITTI & FT & DiffNet \cite{zhou2021self} & 0.7691 $\pm$ 0.0000 & 0.7452 $\pm$ 0.0000 \\
KITTI & Prop. & DiffNet \cite{zhou2021self} & 0.7613 $\pm$ 0.0089 & 0.7821 $\pm$ 0.0117 \\
KITTI & \cite{kuznietsov2021comoda} & DiffNet \cite{zhou2021self} & 0.7892 $\pm$ 0.0141 & 0.8143 $\pm$ 0.0168 \\
KITTI & FT & Prop. & 0.7652 $\pm$ 0.0000 & 0.7296 $\pm$ 0.0000 \\
KITTI & Prop. & Prop. & 0.7866 $\pm$ 0.0016 & 0.8118 $\pm$ 0.0231 \\
KITTI & \cite{kuznietsov2021comoda} & Prop. & 0.7901 $\pm$ 0.0026 & 0.8209 $\pm$ 0.0079 \\
\hline 
vKITTI & FT & DiffNet \cite{zhou2021self} & 0.7962 $\pm$ 0.0000 & 0.7590 $\pm$ 0.0000 \\
vKITTI & Prop. & DiffNet \cite{zhou2021self} & 0.7904 $\pm$ 0.0149 & 0.7685 $\pm$ 0.0101 \\
vKITTI & \cite{kuznietsov2021comoda} & DiffNet \cite{zhou2021self} & 0.7973 $\pm$ 0.0031 & 0.7763 $\pm$ 0.0063 \\
vKITTI & FT & Prop. & 0.7795 $\pm$ 0.0000 & 0.7607 $\pm$ 0.0000 \\
vKITTI & Prop. & Prop. & 0.8057 $\pm$ 0.0058 & 0.7895 $\pm$ 0.0076 \\
vKITTI & \cite{kuznietsov2021comoda} & Prop. & 0.8176 $\pm$ 0.0120 & 0.7855 $\pm$ 0.0046 \\
          \hline
       \end{tabular}
    \end{center}
 \end{table*}

  \begin{table*}
    \caption{$\delta(1.25^2)$ of Stereo-based depth estimation. \textit{FT} stands for fine tuning.}
    \label{tab:kitti_nyu}
    \begin{center}
       \small
       \begin{tabular}{lllcc}
          \hline 
          \begin{tabular}[x]{@{}c@{}}Online\\Train\end{tabular} 
           & Method & Network & TDP & NDP\\ 
          \hline \hline
KITTI & FT & DiffNet \cite{zhou2021self} & 0.9103 $\pm$ 0.0000 & 0.8795 $\pm$ 0.0000 \\
KITTI & Prop. & DiffNet \cite{zhou2021self} & 0.9134 $\pm$ 0.0009 & 0.9030 $\pm$ 0.0078 \\
KITTI & \cite{kuznietsov2021comoda} & DiffNet \cite{zhou2021self} & 0.9187 $\pm$ 0.0020 & 0.9163 $\pm$ 0.0026 \\
KITTI & FT & Prop. & 0.9087 $\pm$ 0.0000 & 0.8806 $\pm$ 0.0000 \\
KITTI & Prop. & Prop. & 0.9182 $\pm$ 0.0005 & 0.9208 $\pm$ 0.0057 \\
KITTI & \cite{kuznietsov2021comoda} & Prop. & 0.9187 $\pm$ 0.0017 & 0.9218 $\pm$ 0.0017 \\
\hline 
vKITTI & FT & DiffNet \cite{zhou2021self} & 0.9152 $\pm$ 0.0000 & 0.8930 $\pm$ 0.0000 \\
vKITTI & Prop. & DiffNet \cite{zhou2021self} & 0.9041 $\pm$ 0.0074 & 0.9088 $\pm$ 0.0042 \\
vKITTI & \cite{kuznietsov2021comoda} & DiffNet \cite{zhou2021self} & 0.9183 $\pm$ 0.0018 & 0.9141 $\pm$ 0.0035 \\
vKITTI & FT & Prop. & 0.9075 $\pm$ 0.0000 & 0.9113 $\pm$ 0.0000 \\
vKITTI & Prop. & Prop. & 0.9194 $\pm$ 0.0059 & 0.9178 $\pm$ 0.0019 \\
vKITTI & \cite{kuznietsov2021comoda} & Prop. & 0.9235 $\pm$ 0.0033 & 0.9178 $\pm$ 0.0013 \\
          \hline
       \end{tabular}
    \end{center}
 \end{table*}

\end{document}